\documentclass[twocolumn]{article}

\DeclareUnicodeCharacter{221E}{\ensuremath{_\infty}}  % workaround for H-inf book title

\usepackage{amsmath, amsfonts}
\usepackage{amssymb}
\usepackage{array}
\usepackage[acronym]{glossaries}
\usepackage{balance}
\usepackage{bm}
\usepackage{booktabs}
\usepackage{cite}
\usepackage{diagbox}
\usepackage{derivative}

\usepackage{enumitem}
\usepackage{epsfig}       % for postscript graphics files
\usepackage{flafter}
\usepackage{graphicx}
\usepackage[hidelinks]{hyperref}
\usepackage{cleveref}
\usepackage{import}
\usepackage[all]{nowidow}
\usepackage{makecell}
\usepackage{mathptmx}
\usepackage{mathtools}
\usepackage{multirow}
\usepackage{ragged2e}     % for \RaggedRight
\usepackage{siunitx}
\usepackage{stfloats}
\usepackage{subcaption}
\usepackage{standalone}
\usepackage{tablefootnote}
\usepackage{tabularx}
\usepackage{textcomp}
\usepackage{verbatim}
\usepackage{wrapfig}
\usepackage{xpatch}% in preamble for line breaks in cells

\makeatletter
\AtBeginDocument{\DeclareMathVersion{bold}
\SetSymbolFont{operators}{bold}{T1}{times}{b}{n}
\SetMathAlphabet{\mathrm}{bold}{T1}{times}{b}{n}
\SetMathAlphabet{\mathit}{bold}{T1}{times}{b}{it}
\SetMathAlphabet{\mathbf}{bold}{T1}{times}{b}{n}
\SetMathAlphabet{\mathtt}{bold}{OT1}{pcr}{b}{n}
\SetSymbolFont{symbols}{bold}{OMS}{cmsy}{b}{n}
\renewcommand\boldmath{\@nomath\boldmath\mathversion{bold}}}
\makeatother

\def\BibTeX{{\rm B\kern-.05em{\sc i\kern-.025em b}\kern-.08em
    T\kern-.1667em\lower.7ex\hbox{E}\kern-.125emX}}
    
\graphicspath{{figures/}}
\newacronym{esu}{ESU}{electrosurgical unit}
\newacronym{pdf}{PDF}{probability density function}
\newacronym{tukf}{TUKF}{truncated unscented Kalman filter}
\newacronym{ekf}{EKF}{extended Kalman filter}
\newacronym{ukf}{UKF}{unscented Kalman filter}
\newacronym{kf}{KF}{Kalman filter}
\newacronym{thermo}{ThERMO}{Thermography for Electrosurgical Rate Modulation via Optimization}
\newacronym{star}{STAR}{Smart Tissue Autonomous Robot}
\newacronym{astr}{ASTR}{Autonomous System for Tumor Resection}
\newacronym{ir}{IR}{infrared}
\newacronym{pva}{PVA}{polyvinyl alcohol}
\newacronym{hpv}{HPV}{human papillomavirus}
\newacronym{occ}{OCC}{oral cavity cancer}
\newacronym{oscc}{OSCC}{oral squamous cell carcinoma}
\newacronym{tors}{TORS}{trans-oral robotic surgery}
\newacronym{lmm}{LMM}{linear mixed model}
\newacronym{he}{H\&E}{hematoxylin and eosin}
\newacronym{iqr}{IQR}{inter-quartile range}
\newacronym{tie}{TIE}{thermal injury extent}
\newacronym{ame}{AME}{absolute margin error}
\newacronym{sme}{SME}{signed margin error}
\newacronym{ldh}{LDH}{lactate dehydrogenase}
\newacronym{nbt}{NBT}{nitro blue tetrazolium}
\newacronym{cdfs}{CDFS}{Clinician-Derived Fixed Speed}
\newacronym{rdfs}{RDFS}{Robot-Derived Fixed Speed}

\DeclareMathAlphabet{\mathcal}{OMS}{cmsy}{m}{n}

\newcommand{\blfootnote}[1]{%
  \begingroup
  \renewcommand\thefootnote{}\footnote{#1}%
  \addtocounter{footnote}{-1}%
  \endgroup
}

\begin{document}

\title{Velocity Optimization via Thermal Imaging Enhances Autonomous Tumor Resection}

\author{Naveed D. Riaziat\footnote{Laboratory for Computational Sensing and Robotics, Johns Hopkins University, Baltimore, Maryland (email: nriazia1@alumni.jh.edu).}
, Jiawei Ge$^*$, Ethan Kilmer$^*$, Michael W. Mikula\footnote{Department of Pathology, Johns Hopkins Hospital, Baltimore, Maryland}
, Paul Maria Scheikl$^*$, \\ Michael Kam$^*$, Leila J. Mady\footnote{Department of Otolaryngology–Head and Neck Surgery, The Johns Hopkins University School of Medicine, Baltimore, Maryland}, Axel Krieger$^*$, and Jeremy D. Brown$^*$ 
}

\maketitle
\blfootnote{Funding for this project was provided by the Link Foundation Modeling, Simulation, and Training Fellowship, ARPA-H grants AY1AX000023 and D24AC00415, and NIH Grant P30CA006973.}
\blfootnote{This work has been submitted to the IEEE for possible publication. Copyright may be transferred without notice, after which this version may no longer be accessible.}

\begin{abstract}
 
\textit{Objective}: To demonstrate and evaluate THERMO, a thermal-velocity optimization algorithm for autonomous electrosurgery, in ex vivo porcine tissue using the UR10e-based ASTR system.
\textit{Methods}: \gls{thermo}, a closed-loop algorithm that uses real-time thermal imaging data to minimize thermal injury while ensuring accurate cut margins, was tested on ex vivo porcine tissue. Histological analysis was performed to detect thermal injury extent at a cellular level, and 3D photogrammetric measurement was used to verify margin accuracy. \gls{thermo}'s performance was compared against two previously used fixed velocity methods.
Results: \gls{thermo} outperformed both fixed velocity comparison methods without requiring data-driven training or iterative tuning, reducing thermal injury by up to 52\%, improving cut accuracy by up to 59\%, and maintaining task completion speed.
\textit{Conclusions}: Closed-loop thermal feedback via \gls{thermo} improves autonomous robotic electrosurgery by simultaneously reducing thermal injury and improving margin accuracy, reinforcing its value over fixed velocity approaches for tumor resection applications.
\textit{Significance}: This work advances autonomous robotic tumor resection by demonstrating that real-time closed-loop thermal feedback control can substantially reduce collateral tissue damage and improve surgical margin accuracy.
\end{abstract}

%%%%%%%%%%%%%%%%%%%%%%%%%%%%%%%%%%%%%%%%%%%%%%%%%%%%%%%%%%%%%%%%%%%%%%%%%%%%%%

\section{Introduction}

% \csname documentclass\endcsname[...]{subfiles}
% \graphicspath{{\subfix{../figures/}}}
% \begin{document}

\begin{figure}[t]
    \centering
    \includegraphics[width=\linewidth]{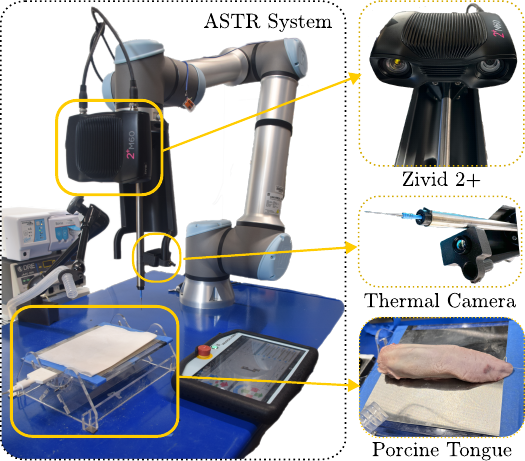}
    \caption{The \acrshort{astr} system is an autonomous robotic surgical system for conducting autonomous electrosurgical tumor resection on ex vivo tissue. It consists of a UR10e arm (Universal Robots, Odense, Denmark), a Zivid 2+ RGB-D camera (Zivid, Oslo, Norway), and an electrosurgical system. A  T3 Pro thermal camera (InfiRay, Yantai, China) is attached to the tool flange to provide thermal feedback for \gls{thermo}. The porcine tongue sample is placed on the grounding pad on the table and the UR10e autonomously performs tumor resection using information from the Zivid and T3 cameras.}
    \label{fig:astr}
\end{figure}

\Gls{occ} is cancer of the mouth, tongue, lips, and cheeks. Approximately 377,000 new \gls{occ} cases are diagnosed each year globally, leading to 178,000 deaths. Tobacco use, alcohol consumption, and \gls{hpv} are risk factors for developing \gls{occ}. Over 90\% of \gls{occ} cases are squamous cell carcinoma, or cancer of the cells that line the outer layer of the tongue, cheeks, gums, and throat~\cite{kain_surgical_2020}. For treatment of pre-metastatic \gls{oscc}, surgical intervention is often the most effective treatment~\cite{noauthor_tongue_2022}. 

The standard surgical approach is the partial glossectomy procedure, where the portion of the tongue consisting of the \gls{oscc} tumor is removed. Cancer recurrence rates are minimized if the surgical margins (the region between the tumor and the cut boundary) reach \SI{5}{\milli\meter}~\cite{seethala_data_2019}. The term negative margin indicates that no cancer cells were found in the margin, while positive margin indicates the opposite. Though positive margins increase recurrence risk, excess negative margins indicate excess healthy tissue removal and worsen post-operative recovery and quality of life~\cite{eldeeb_effect_2012, li_margin_2019}. Therefore, it is critical to ensure accurate surgical margins during tumor resection. 

Electrosurgery is a common technique for removing tumor tissue, wherein a surgical tool conducts electricity to generate heat in tissue, thereby causing it to denature. The heat-induced tissue denaturation reduces mechanical cutting effort while controlling bleeding. Monopolar electrosurgery directly heats the tissue by passing current from the tool tip through the body to an alternate-site grounding pad. Excess application of electrosurgery, however, may cause trauma to the lingual nerve in the tongue and permanently impair tongue sensory function. Insufficient application, by contrast, reduces cut accuracy by causing tissue snagging, which negatively impacts surgical margins and outcomes. As such, electrosurgery requires surgeons to diligently monitor tissue conditions to prevent adverse outcomes.

Monitoring and safely controlling electrosurgery becomes increasingly difficult in the tight constraints of the oral cavity. Consequently, robotics has provided advances in ergonomics and visualization during oral surgeries~\cite{bhayani_shifting_2010, chung_transoral_2015}. Systems like the da Vinci Single Port (Intuitive Surgical Inc., Sunnyvale, California, USA) have made \gls{tors} possible by reducing tool size while maintaining high-quality stereoscopic vision. The application of robotics to head and neck surgery offers the possibility of procedural enhancements with sensing, feedback, and automation.

Though current commercial soft tissue surgical robots have little to no autonomous capability, increased electrosurgical autonomy could greatly reduce surgeon workload and improve surgical outcomes~\cite{fiorini_concepts_2022}. In the few early demonstrations of autonomous tumor resection has been an active research field, a discrepancy exists between the surgeon's technique and the robot's. While experienced surgeons adjust their cutting speed based on visual feedback of cut accuracy and thermal injury, current autonomous electrosurgical robots do not. 

\subsection{Related Work}

Autonomous surgical debridement of bone tissue was first demonstrated in human cadavers by Sefati et al. \cite{sefati_dexterous_2022}. They used a hybrid continuum robot on a UR10 robot to perform osteolytic debridement in the pelvis using a surgeon-approved preoperative plan. Autonomous cutting of hard tissue like bone does not face many of the challenges of cutting soft tissue. Autonomous surgical cutting has typically run in an open-loop fashion, using a velocity predetermined via trial-and-error tuning or simulation. Studies have shown that autonomous robotic electrosurgery of this form causes more cellular injury than manual electrosurgery, indicating that the fine adjustments surgeons make during surgery have a substantial impact on excess heat injury~\cite{ewertowska_histological_2024}. In past demonstrations of autonomous tumor resection, researchers have prioritized cutting accuracy over thermal injury and manually selected low cutting speeds. For instance, Opfermann et al.\ first demonstrated a visual servoing approach for electrosurgical cutting on the \gls{star}~\cite{opfermann_semi-autonomous_2017, shademan_supervised_2016}. Saeidi et al.\ similarly showed that STAR could perform tissue-cutting tasks using a predetermined cut speed~\cite{saeidi_supervised_2019}. Ge et al.\ introduced the \acrfull{astr} system for fully autonomous tumor resection and demonstrated successful cutting at \SI{2}{\milli\meter\per\second}~\cite{ge_autonomous_2024}. Ge et al.\ later proposed that for faster velocities that excessively deform tissue, simulation-based trajectory augmentation could reduce spatial error~\cite{ge_enhancing_2024}. However, soft tissue simulation is both challenging and not patient-specific, potentially limiting robustness, optimality, and safety without additional feedback mechanisms. 

Closed-loop sensing has shown promise in optimizing electrosurgical performance. Since thermal injury is largely invisible to the naked eye until permanent damage has been done, researchers have used thermal imaging to measure electrosurgical thermal spread~\cite{nechay_thermal_2020, wallwiener_quantifying_2010, siracusano_measuring_2023}. Recent works have developed miniaturized endoscopic thermal cameras as well, which are compatible with the laparoscopic surgery workflow~\cite{collins_thermal_2019, matsuura_flexible_2011, scutelnic_thermal_2022}. Bao and Mazumder showed that thermal imaging can control the denaturation zone using a novel computer-controllable \gls{esu}~\cite{bao_reduced_2022}. Surgeons, however, typically employ a fixed power level, and most \glspl{esu} are not capable of real-time power modulation. El-Kebir et al.\ instead used thermal sensing to train a data-driven non-linear autoregressive exogenous input (NLARX) model to determine when a static surgical tool should move to the next waypoint~\cite{el-kebir_minimally_2023}. It is still unclear, however, if these data-driven models generalize from laboratory settings to the complex surgical environment. Some researchers instead have developed methods to estimate soft tissue parameters online, including by use of thermography, Bayesian and Gaussian Estimation, and finite element models~\cite{shin_new_2016, hossain_tumor_2016, zhu_iterative_2023, chalasani_preliminary_2018, noe_gaussian_2019}. 

While these prior works have demonstrated the potential benefits of thermal injury measurement and control, electrosurgery is more appropriately solved as an optimal control problem with competing goals of limiting thermal injury and maximizing cut accuracy, which is significantly impacted by tissue deformation. In an attempt to maximize accuracy, Shademan et al.\ used near-infrared fluorescent markers to track tissue deformation during autonomous porcine intestinal anastomosis~\cite{shademan_supervised_2016}. Kam et al.\ showed promising results controlling electrosurgical cut velocity with tracker-based tissue deformation feedback~\cite{kam_autonomous_2025}. Similarly, Ferrari et al.\ used deformation modeling and feedback to track and therefore more accurately target tumor biopsy sites~\cite{ferrari_autonomous_2023}. Giannarou et al.\ utilized microscopy images during retinal surgery to estimate tissue deformation and tool forces~\cite{giannarou_vision-based_2016}. Despite these advancements, the combination of deflection and thermal injury feedback has yet to be demonstrated. Without considering both deflection and thermal injury, the above approaches to autonomous electrosurgical control cannot achieve optimality, especially in the highly muscular and therefore tough tissue found in the tongue. 

\subsection{Contributions}

Our previous work demonstrated optimal autonomous electrosurgical cutting speed in tissue phantoms using a thermal-feedback algorithm, \acrfull{thermo}. \gls{thermo} improved resistance to disturbances, reduced thermal injury, and minimized tool deflection without the need for simulation or model training~\cite{riaziat_towards_2025}. \Gls{thermo} prevented the need for trial-and-error speed selection by using online tissue parameter identification. However, up to this point, autonomous cutting under visual thermal feedback has not yet been demonstrated for 3D cutting in biological tissues with a multi-DoF robotic system. Testing in ex vivo tissue is a critical step towards eventual cadaveric and in vivo studies. Likewise, the connection between thermal feedback and cell death and the relationship between tool deflection and margin accuracy have not been validated through histological analysis or post-hoc margin measurement. Validating these outcomes is critical for the eventual clinical translation of autonomous thermal-feedback electrosurgical technologies.

Here, we evaluate adaptive velocity control using \gls{thermo} on the \acrfull{astr} system, an autonomous robotic surgery platform with an electrosurgical tool and RGB-D camera (see Fig.~\ref{fig:astr})~\cite{ge_autonomous_2024}. Our contributions are threefold. First, we validate \gls{thermo}'s effectiveness in improving margin accuracy while minimizing thermal spread in an ex vivo porcine tissue model. Second, we present histological analysis quantifying thermal injury in relation to cutting accuracy during a simulated electrosurgical tumor resection. Third, we provide a comparative study measuring thermal injury and cut accuracy against fixed-speed robotic methods, highlighting the benefits of thermal feedback in surgical performance. These results underscore the promise of intelligent electrosurgical automation for improving tumor resection outcomes in future surgeries.

% \end{document}

\section{Methods and Materials}

% \csname documentclass\endcsname[...]{subfiles}
% \graphicspath{{\subfix{../figures/}}}

% \begin{document}
\subsection{Current Clinical Resection Technique}

Most \gls{oscc} occurs on the tongue, where the mortality rate is also the highest~\cite{almangush_staging_2020}. \Gls{oscc} tumors are typically \SIrange{2}{4}{\centi\meter} in diameter and the depth of invasion is typically $\leq$ \SI{10}{\milli\meter} when surgical intervention is most effective~\cite{almangush_staging_2020}. A common surgical technique for tumor resection is called a ``partial glossectomy", where a portion of the tongue containing the tumor is removed.

In cases where the cancerous tissue is small and near the tip of the tongue, access to the tumor may be obtained trans-orally. For larger tumors near the base of the tongue which cannot be accessed easily, a trans-cervical pull-through or mandibulotomy may be performed to create additional space to operate. Robotic approaches often have better small-space dexterity and access and thus can avoid additional access procedures. With access gained and the tongue prepared, the tongue is sutured at its tip to pull it from the oral cavity and provide tension on the tissue. Then, the tumor outline is marked and the cutting boundary is defined \SI{5}{\milli\meter} from this edge~\cite{park_first_2019}. This space between the tumor and the cut defines the resection margin, and is critical to ensuring no cancerous cells are left behind. Then, with the goal of precisely maintaining this margin to prevent excess tissue damage or leftover cancerous tissue, the surgeon will use an electrosurgical tool to cut around this defined periphery, eventually approaching from the underside of the tumor to complete the resection. The removed samples will be sent to pathology for margin confirmation; if the margins are positive (there is no healthy tissue around the tumor), the surgeon will remove additional tissue until there are negative margins. The studies presented here will focus on the first peripheral cut step of the above, as they are most critical in determining surgical margins.

\subsection{Autonomous System for Tumor Resection} \label{ssec:astr} 
% Robotic testbed. ROS-based low and mid-level control.

The \gls{astr} system, shown in Fig.~\ref{fig:astr}, is a software and hardware system consisting of a monopolar electrosurgical tool and Zivid 2+ structured-light RGB-D camera (Zivid, Oslo, Norway) mounted to a UR10e robot~\cite{ge_autonomous_2024}. The UR10e arm is controlled using the Open Robot Control Software Kinematics and Dynamics Library and ROS 2 on Ubuntu 22.04. Trajectory generation is computed using the Reflexxes Motion Library.

The Zivid 2+ is used to measure the tissue surface with \SI{55}{\micro\meter} accuracy for trajectory planning. The tool and grounding pad are connected to the \gls{esu}, which is the source of electrosurgical energy. The tool is the electrical anode, and the grounding pad, typically attached to the patient's back, serves to close the current loop. The \gls{esu} generates high-frequency current with selectable wattage and waveform shapes. Different waveforms allow for different effects, such as coagulation and fulguration. Typical wattage varies between \SIrange{20}{50}{\watt}. For these experiments, the \gls{esu} is set to \SI{35}{\watt} and cut mode as advised by an expert head-and-neck surgeon. 

\subsection{ThERMO}

\Gls{thermo} is an online parameter estimation and optimization framework that uses thermal imaging to determine the optimal electrosurgical cutting velocity via a prototypical thermal and mechanical model of the tissue with unknown parameters. Here, a brief overview is provided; for a more detailed discussion, please refer to our previous work.~\cite{riaziat_towards_2025}. In the preceding work, we developed the thermal sensing system, adaptive parameter identification, and optimization techniques underlying the thermo device. Results were demonstrated in 2D hydrogel phantoms, showcasing the potential of extending the work to 3D cuts in biological tissue. The iteration of \gls{thermo} used in this article is minimally modified from the previous work, with only necessary adjustments for the new robot’s kinematic constraints. This should further demonstrate that manual tuning is not required to achieve satisfactory performance with the thermo device.

\subsubsection{Deformation and Thermal Injury Models}

A model is developed to predict the thermal injury extent by extracting the width of the denaturation isotherm. The denaturation isotherm is defined at \SI{60}{\celsius}~\cite{sapareto_thermal_1984}. The electrosurgical cut is modeled as a 2D moving heat source problem, since the cuts are shallow and the curvature is low. The thermal model from the 2D heat equation is thus

\begin{equation}
    \pdv{T}{t} = \alpha ( \pdv[order=2]{T}{x} + \pdv[order=2]{T}{y}),
\end{equation}

\noindent where $\alpha$ is the thermal diffusivity, $T$ is the temperature field, and $t$ is time. It is assumed there is no heat transfer via radiation or convection, only conduction. Lu et al.\ developed a model to predict isotherm widths~\cite{lu_width_2020} from a quasistatic approximation of the 2D heat equation~\cite{weichert_temperature_1978}. The selected isotherm temperature $T_c$ is the denaturation temperature referenced above at \SI{60}{\celsius}. The asymptotic isotherm width $w$ with correcting factor $f(T^*_c)$ is given as a function of the control velocity $u$: 

\begin{gather}  
    w(u) = \frac{1}{\sqrt{2\pi e}}\frac{q \alpha}{u \lambda d (T_c - T_o)}\,f(T^*_c), \label{eq:thermal_width} \\
    f(T^*_c) \approx \exp(-1/T^*_c)[1 + (1.477T^*_c)^{1.407}]^{0.7107}
\end{gather}

\noindent where the unit-less isotherm temperature $T^*_c = 2\pi\,\lambda\,d\,(T_c - T_0)\,q^{-1}$, $T_c$ is the isotherm temperature, $T_o$ is the ambient tissue temperature, and $\gamma = 0.5772$ is the Euler-Mascheroni constant. 

Cutting force contributes to reduced cut accuracy due to tissue snagging and deformation. Due to Hooke's law, tool deformation can be used as a proxy for directly measuring force. The cutting force as a function of velocity, adjusted to match empirical observations, is:

\begin{equation}
    F_{cut}(u) = -d\,\exp(-C_{\text{defl}}/u) ,
    \label{eq:cutting_force}
\end{equation}

\noindent such that $d$ encodes the maximum possible force which grows at a rate $C_{\text{defl}}$ with respect to velocity~\cite{bain_thermomechanical_2011}.

% The discrete second order dynamics at time step $i$ thus are

% \begin{align}
%     x_{i} &= \begin{bmatrix}
%         x_{t, i} & \dot{x}_{t, i} & x_{n, i} & \hat{C}_{\text{defl}}
%     \end{bmatrix}^T,  \label{eq:discrete_tip_dynamics_state} \\ 
%     x_{i+1} &= f(x_i, u_i) + \mathcal{N}(0, Q),  \label{eq:discrete_tip_dynamics_predict} \\
%     z_{i} &= \begin{bmatrix}
%         x_{t, i} & x_{n, i}
%     \end{bmatrix}^T + \mathcal{N}(0, R), \label{eq:discrete_tip_dynamics_measure} 
% \end{align}

% \noindent \noindent where $x_t \in \mathbb{R}^2$ is the tool position, $k$ is the tool mechanical stiffness, $x_n \in \mathbb{R}^2$ is the tool position when $|F_{cut}| = 0$, and the state transition function $f(x_i, u)$ is 

% \begin{equation}
%     f(x_i, u) = x_i + \Delta\,t\begin{bmatrix}
%          \dot{x}_{t,i} \\
%          -k (x_{t, i} - x_{n, i}) + F_{cut} \\
%          \vec{0} \\ 
%     \end{bmatrix}.
% \end{equation}

\begin{figure}[tbp]
    \centering
    \includegraphics[width=\linewidth]{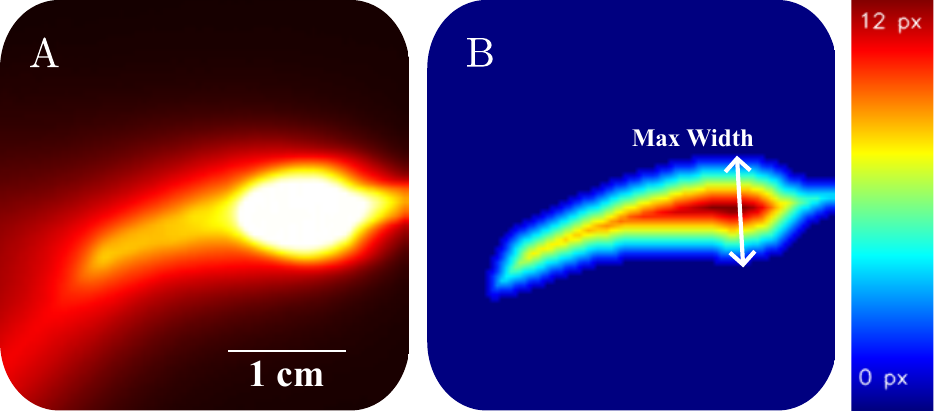}
    \caption{(A) A thermal frame is shown from the T3 thermal camera, looking down the shaft of the electrocautery needle. The white location corresponds to tissue contact from the needle, and the hot trail shows the cutting path. (B) The L2 distance transform is shown, in which the color corresponds to the distance a given pixel is from the \SI{60}{\celsius} temperature boundary. The thermal width is the maximum value of this array.}
    \label{fig:cv}
\end{figure}

\subsubsection{Thermal Imaging} 

\Gls{thermo} receives thermal images from a uncooled bolometer type thermal camera, the T3 Pro (IRay Technology Co., Ltd., Shanghai, China). It captures a 384 x 288 pixel array at \SI{25}{\hertz} with \SI{2}{\celsius} or 2\% accuracy, whichever is greater. Example thermal images are shown in Fig.~\ref{fig:cv}A and Fig.~\ref{fig:timeline}. Spatial, temporal, and thermal accuracy is confirmed as shown in~\cite{riaziat_towards_2025}. The camera is mounted to the robot end effector flange so that the tool frame is rigidly attached to the camera frame.  Two measurements are deduced using the thermal camera: 1) isotherm width and 2) tool deflection. To measure the isotherm width, the thermal frames are thresholded at \SI{60}{\celsius}. Then, an L2 distance transform is performed, creating an image where the value of each pixel is the Euclidean distance to the nearest point under \SI{60}{\celsius}. The maximum of the distance transform corresponds to half the thermal width, as shown in Fig.~\ref{fig:cv}. The tool deflection is measured by extracting the hottest point in the image, which corresponds to the contact location between the electrosurgical tool and the tissue. The motion of this point relative to its resting position defines the deflection.

\subsubsection{Parameter Estimation}

Dual state and tissue parameter estimation is conducted using the \gls{tukf}. The \gls{ukf} is a Kalman filter for non-linear models that uses the ``Unscented Transform" in place of Jacobian computation from the more common \gls{ekf}. The \gls{ukf} estimates $\hat{\lambda}$, the augmented tissue thermal conductivity, $\hat{d}$, the augmented maximum cutting force, $\hat{q}$, the linear power density, and $\hat{c}$, the tissue specific heat capacity. Unknown parameters are initialized from values published by the Foundation for Research on Information Technologies in Society (IT'IS)~\cite{itis_foundation_tissue_2024}, shown in Table~\ref{tab:combined_params}.

\subsubsection{Optimization}

Successful electrosurgical cutting relies on balancing cutting speed between moving too fast and too slow. If the cut is performed too quickly, little electrical heat will be generated, and tissue will not denature enough to reduce the cutting force. As a result, tissue will snag, and cut accuracy will decrease. If the cut is performed too slowly, excess heat will injure healthy tissue, causing unnecessary injury to the patient and worsening post-operative outcomes. Thus, a natural minimization problem arises between cutting force and thermal injury. This optimization is covered in detail in our previous work so only a brief summary is given here~\cite{riaziat_towards_2025}. Consider, at a given time step $i$, the following cost function:

\begin{multline}
    J_i(v_i) = a F_i(v_i)^2 + b w_i(v_i)^2 + c (\min(v_i - \bar{v}, 0)) \\ + r (v_i - v_{i-1})^2, 
    \label{eq:cost_func}
\end{multline}

\noindent where $v_i$ is the tool velocity, $F_i$ is the cutting force, $w_i$ is the width of thermal injury, $\bar{v}$ is a nominal velocity estimate, and $a, b, c, r$ are weights. For these experiments, the selected weights are shown in Table~\ref{tab:combined_params}. Weights $a$ and $b$ were selected to reflect equal clinical importance between thermal and mechanical damage, while weights $c$ and $r$ were tuned by setting the minimum value that resulted in improved motion smoothness. 

\begin{table}[tp]
    \centering
    \small
    \caption{Thermal properties: $\rho$ (tissue density), $C_p$ (specific heat), $k$ (thermal conductivity), $\alpha$ (thermal diffusivity). Thermal properties and cost function weights $a,b,c,r$ as defined in Eq.~\ref{eq:cost_func}.}
    \label{tab:combined_params}
    \begin{tabular}{ll ll}
        \toprule
        \multicolumn{2}{c}{\textbf{Thermal Properties}} & \multicolumn{2}{c}{\textbf{Cost Function Weights}} \\
        \cmidrule(lr){1-2} \cmidrule(lr){3-4}
        $\rho$ & $1.09\times10^{-6}$ kg/mm$^3$ & $a$ & 1 \\
        $C_p$ & 3800 J/kg/K & $b$ & 1 \\
        $k$ & $5.6\times10^{-4}$ W/mm/K & $c$ & 5 \\
        $\alpha$ & $1.34\times10^{-7}$ mm$^2$/s & $r$ & 0.01 \\
        \bottomrule
    \end{tabular}
    \vspace{1.0em}
\end{table}

Optimization is performed using \texttt{minimize\_scalar} from SciPy's optimize library in Python 3.10, which uses Brent's method to solve scalar minimization quickly. As the cost function is scalar and quadratic, only one minima exists, and it can be found rapidly (fraction of a millisecond) by almost any solver. Note this is orders of magnitude faster than our thermal sensing's 30\,Hz sampling rate.

\begin{figure}[tb!]
    \centering
    \includegraphics[width=\linewidth]{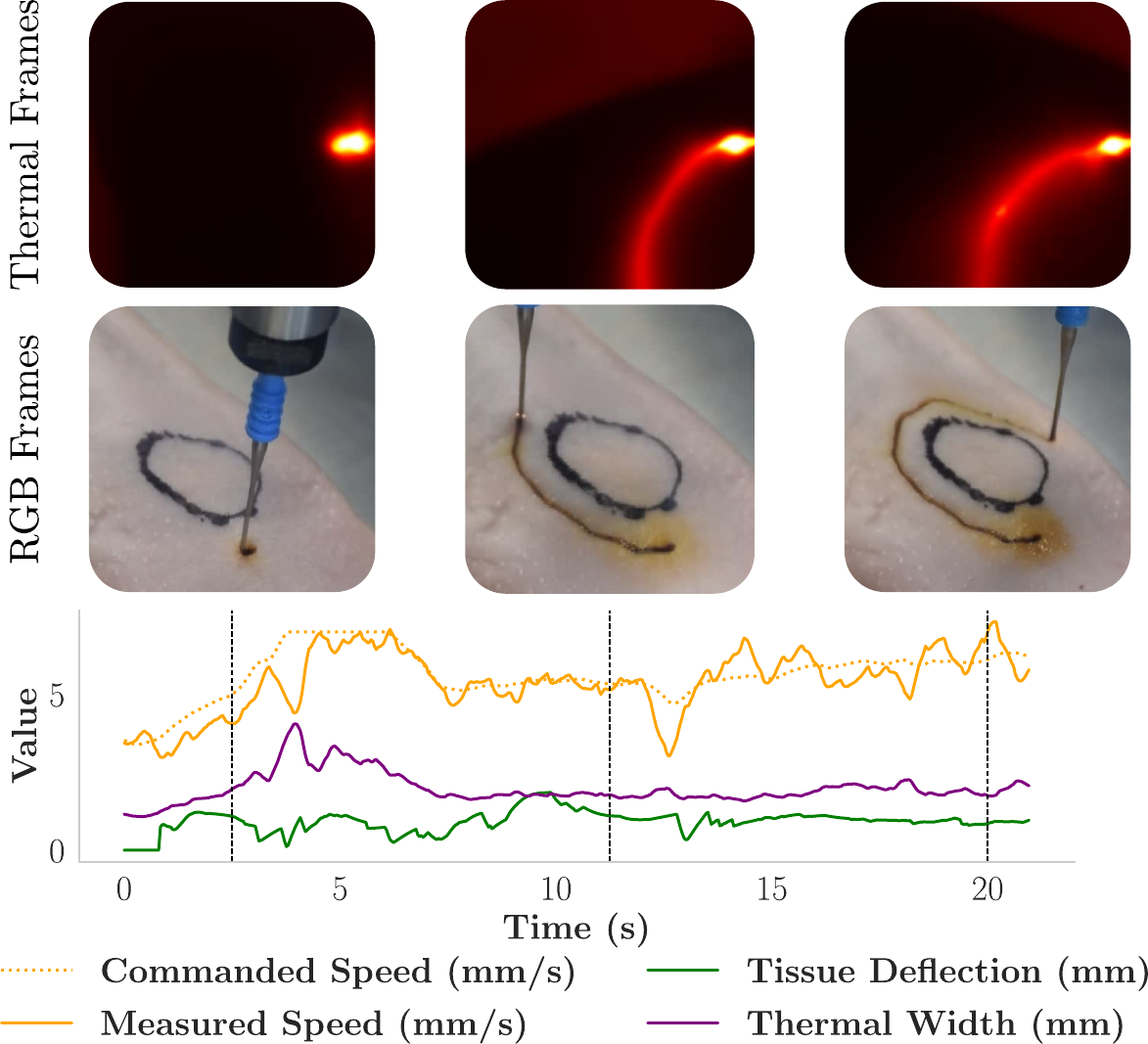}
    \caption{Commanded velocity, measured velocity, thermal width, and deflection are plotted with respect to time for a selected trial of \gls{thermo}. Thermal and RGB frames are displayed at three timestamps above the plot, showing the correlation between the visible and thermal cutting states.}
    \label{fig:timeline}
\end{figure}

\subsection{Trajectory Planning}

\begin{figure}[tp]
    \centering
    \includegraphics[width=\linewidth]{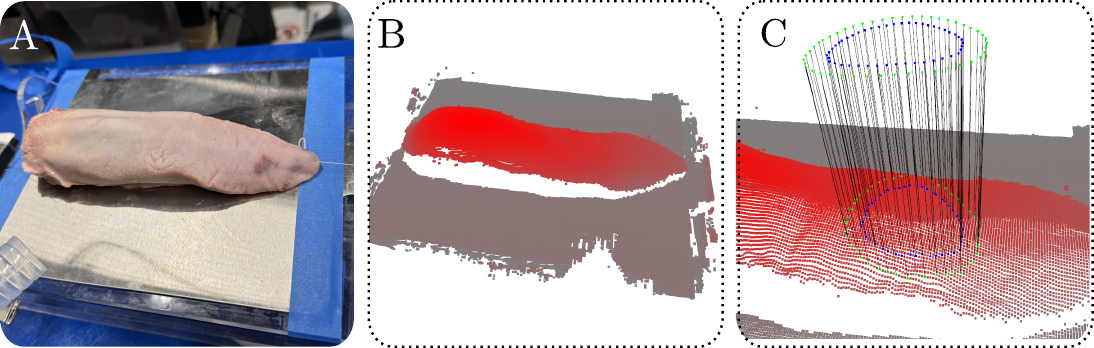}
    \caption{(A) Tongues are placed on an electrosurgical grounding pad and sutured as in the partial glossectomy procedure to prevent bulk tissue motion. (B) Pointcloud scan from the Zivid 2+ camera. (C) Projection of tumor outline onto pointcloud.}
    \label{fig:tongue_pcd}
\end{figure}

Several constraints must be implemented to define a cutting trajectory on a tissue sample. All cuts are performed with a prototypical tumor outline determined from a real clinical tongue tumor scan (see Figs.~\ref{fig:timeline}). Two trajectories are generated from this outline: the first defines the boundary of the tumor itself and the second defines the targeted \SI{5}{\milli\meter} cutting margin extended from this boundary. The following procedure is then conducted to plan the robot trajectory (see Fig. \ref{fig:tongue_pcd}). 

\begin{enumerate}
    \item The tongue is fixed to the grounding pad using surgical suture. 
    \item A 3D point cloud is captured using the Zivid 2+. 
    \item The boundary of the tumor and the \SI{5}{\milli\meter} cut margin are projected onto the tissue surface by manual selection of a central region of the anterior tongue. A discrete set of points and normal vectors are output for each trajectory. 
    \item The tangent vectors between subsequent trajectory points are computed. The tangent vectors are projected onto the local surface. 
    \item The full pose at each point is determined by orienting the tool x-axis with the projected vector and the tool z-axis with the negative of the surface normal. 
    \item The trajectory orientations are smoothed using an exponential filter. 
\end{enumerate}

After each 30~Hz imaging update, the newly computed desired velocity ($<1~ms$ computation time) is passed to an Orocos Real-Time Toolkit (RTT) component running at 125~Hz on Ubuntu. This component generates smooth joint-space trajectory setpoints using the Reflexxes motion library and publishes them to the Universal Robots ROS~2 driver, which streams setpoints to the UR controller over RTDE (Real-Time Data Exchange) at 125~Hz. This low-level layer performs deterministic trajectory generation and tracking without iterative optimization. The optimization resides solely in the $\sim$ 30~Hz application-specific high-level planner, which is not considered real-time.

The tool x-axis must be oriented in the direction of motion such that the thermal camera trails the cutting edge. This ensures the tool's thermal ``wave" is always in frame (see Fig.~\ref{fig:timeline}). Second, the tool z-axis should be oriented normal to the surface of the tissue to minimize any perspective warping of the isotherm size. This procedure fully defines the poses for the trajectory. The command velocity for each pose is either set a priori or adjusted in real time during a cut based on the cutting method.

\subsection{Experimental Procedure}

\begin{figure*}[tp]
    \centering
    \includegraphics[width=\linewidth]{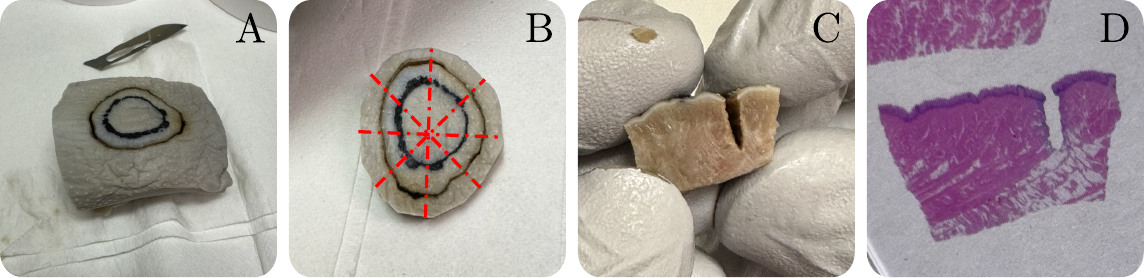}
    \caption{(A) Tongue sample after 72 hours of formalin fixation. (B) Tongue sample cut to rough size, with red dashed lines indicating how each sample was cut into eight wedge-shaped blocks that evenly sampled the electrosurgical cut spatially. (C) A cross-section of one of the wedge-shaped blocks resulting from the previous step. This face is cut on a microtome to generate three \SI{5}{\micro\meter} thick sections from each block. (D) A slide with a \gls{he} stained and mounted tissue section.}
    \label{fig:sectioning}
\end{figure*}

Ten porcine tongues were procured from Animal Technologies\footnote{https://www.animaltechnologies.com}, as porcine tongues have similar properties to human tongues~\cite{stembirek_pig_2012}. Packaged samples were airmailed in ice water overnight after sacrifice, and experiments were conducted within 36 hours of receipt. After each trial, samples were immediately cut to size and fixed in 10\% formalin for 72 hours (see Fig. \ref{fig:sectioning}).

Experiments were conducted in three groups for each of the three different cutting methods. Each group executed three cuts on separate tongues. One tongue was not cut and used as a histological control. As per the most recent works on autonomous tongue tumor resection, one group is run at \SI{2}{\milli\meter\per\second}. Then, based on measurements of handheld electrosurgical cutting from an expert surgeon, a second group is cut at \SI{7}{\milli\meter\per\second}~\cite{ge_enhancing_2024}. The final group utilizes the \gls{thermo} method to optimize cut velocity. The three methods are thus referred to as follows:

\begin{itemize}
    \item \textbf{\protect{\gls{rdfs}}}: \SI{2}{\milli\meter\per\second}, as previously used in the \gls{astr} system~\cite{ge_autonomous_2024},
    \item \textbf{\protect{\gls{cdfs}}}: \SI{7}{\milli\meter\per\second}, as measured by external trackers on an expert surgeon's handheld electrosurgical device~\cite{ge_enhancing_2024},
    \item \textbf{\protect{\gls{thermo}}}: adaptive thermal-feedback optimized velocity.
\end{itemize} 

These values are also summarized in Table~\ref{tab:cutting_methods_and_metrics}.

\subsection{Measurements}

Four quantitative measures are reported and described below: 1) \textbf{\protect{\gls{sme}}}: error between the measured and desired margin, 2) \textbf{\protect{\gls{ame}}}: absolute error between the measured and desired margin, 3) \textbf{\protect{\gls{tie}}}: average perpendicular distance of thermal injury from the cut, 4) task completion time. These measures are also summarized in Table~\ref{tab:cutting_methods_and_metrics}. 

\subsubsection{Cut Accuracy}

Before each margin cut, the boundary of the virtual tumor is marked onto the surface using an ink pen attached to the robot arm in place of the electrosurgical tool. After cutting, the distance between the two contours is photogrammetrically measured by an Intel RealSense D405 camera (Intel Corporation, Santa Clara, California, USA). The D405 was used in place of the Zivid to ensure unbiased and reliable result validation by using an independent sensor. The D405 returns a color image registered to a point cloud. The cut boundary and the tumor margin are segmented from the color image and correlated to their 3D points. Points in the inner and outer contours are correlated by angle, and the 3D Euclidean distance is computed. \gls{ame} and \gls{sme} are reported, representing both the absolute accuracy and the clinically-relevant signed accuracy. 

Additionally cutting depth was also measured under microscopy to ensure verify accuracy relative to the desired \SI{5}{\milli\meter} cut.

\subsubsection{Thermal Injury}

Histological analysis is used to determine the extent of thermal injury caused by the electrosurgery instrument. From each tongue, eight wedge-shaped blocks of tissue are excised orthogonally to the cut direction. These blocks are fixed in formalin, paraffin-embedded, and sectioned using a microtome into three \SI{5}{\micro\meter} sections at \SI{50}{\micro\meter} increments. Sections are stained with \acrfull{he}, a standard stain used for microscopy, and mounted to slides. A visual guide to the sectioning process is shown in Fig.~\ref{fig:sectioning}. A board-certified pathologist, blind to the cutting method, then analyzes the slides for thermal injury and reports the amount by which thermal injury extends from the cut location. From each slide, eight perpendicular measurements are taken around the cut to determine the distance of thermal injury away from the cut. The average of these eight values is reported as \gls{tie}. \Gls{he} stained samples highlight nuclear and cytoplasmic structures, enabling pathologists to evaluate tissue cytoarchitecture (i.e., the structural organization of cells) and assess thermal injury. Previously characterized morphological indicators of thermal damage range from connective tissue bluing and smudging of nuclei to complete loss of cytoarchitecture~\cite{nagarajan_association_2020, ruidiaz_quantitative_2011}.

\subsubsection{Task Completion Time} 

Task completion time is defined as the time it takes from the moment the electrosurgical tool makes contact with the tissue to the moment it breaks contact. Since time is a valuable resource in the operating room, this metric is key in determining efficiency and optimality for clinical translation. To note, the reported time reflects only the duration of the tongue surface incision, which is a subtask within the broader tongue tumor resection surgery. It does not represent the total procedure time.

\subsection{Statistical Analyses}

Statistical analyses of the metrics 1) \gls{ame}, 2) \gls{sme}, and 3) \gls{tie} (described above) are conducted with \glspl{lmm}, a linear-regression-based statistical analysis that handles nested study structures accurately. \glspl{lmm} use fixed effects, such as treatment method, and random effects, such as trial, to robustly infer statistical results with repeated, hierarchical, and correlated data, unlike ANOVA, which assumes independence of observations and does not natively account for within-subject or within-trial variability. The models for \gls{ame} and \gls{sme} use cutting method as a fixed predictor and the trial as a random predictor, thereby accounting for biological variations in tongues. The model for \gls{tie} adds a random predictor of the cut block to account for the repeated measures within a cross-section.

% Here, the following models are used:

% \begin{equation}
%     \begin{aligned}
%     \text{AME} &\sim \text{Method} + (1\,|\,\text{Trial}) \\
%     \text{SME} &\sim \text{Method} + (1\,|\,\text{Trial}) \\
%     \text{TIE} &\sim \text{Method} + \text{Depth} + (1\,|\,\text{Trial}) + (1\,|\,\text{Block})
%     \end{aligned}
% \end{equation}

Method is a categorical variable referring to either \gls{rdfs}, \gls{cdfs}, or \gls{thermo}. Depth is a quantitative variable measuring the depth of cut seen in the histology slides. Both are used as fixed effects to determine their direct relationship on the outcome variables. Trial and Block are categorical variables referring to the trial number (and therefore the tongue being cut) and the wedge-shaped tissue block shown in Fig.~\ref{fig:sectioning}C, respectively. They are included as a random effect to indicate biological differences between tongues and to account for the repeated measurement between the three sections removed from the same block. For all models, visual inspection confirmed normality of residuals, homoscedasticity, and homogeneity of variances. Results are reported with $p$-values and standard errors. 

\subsection{Qualitative Clinical Analysis}

A board-certified expert otolaryngologist, blinded to the experimental method, evaluated the three electrosurgical cutting methods using unlabeled image–video pairs corresponding to a cut performed on porcine tongues. Each method, labeled Method 1, Method 2, and Method 3, was rated on a 5-point scale (1-poor, 5-excellent) based on margin accuracy, thermal effects, and overall impression. These measures are intended to supplement the other quantitative results and provide clinical insight that would otherwise not be communicated.

% \end{document}

\section{Results}

% % \csname documentclass\endcsname[...]{subfiles}
% % \graphicspath{{\subfix{../figures/}}}

% \begin{document}

\begin{figure*}
    \centering
    \includegraphics[width=\linewidth]{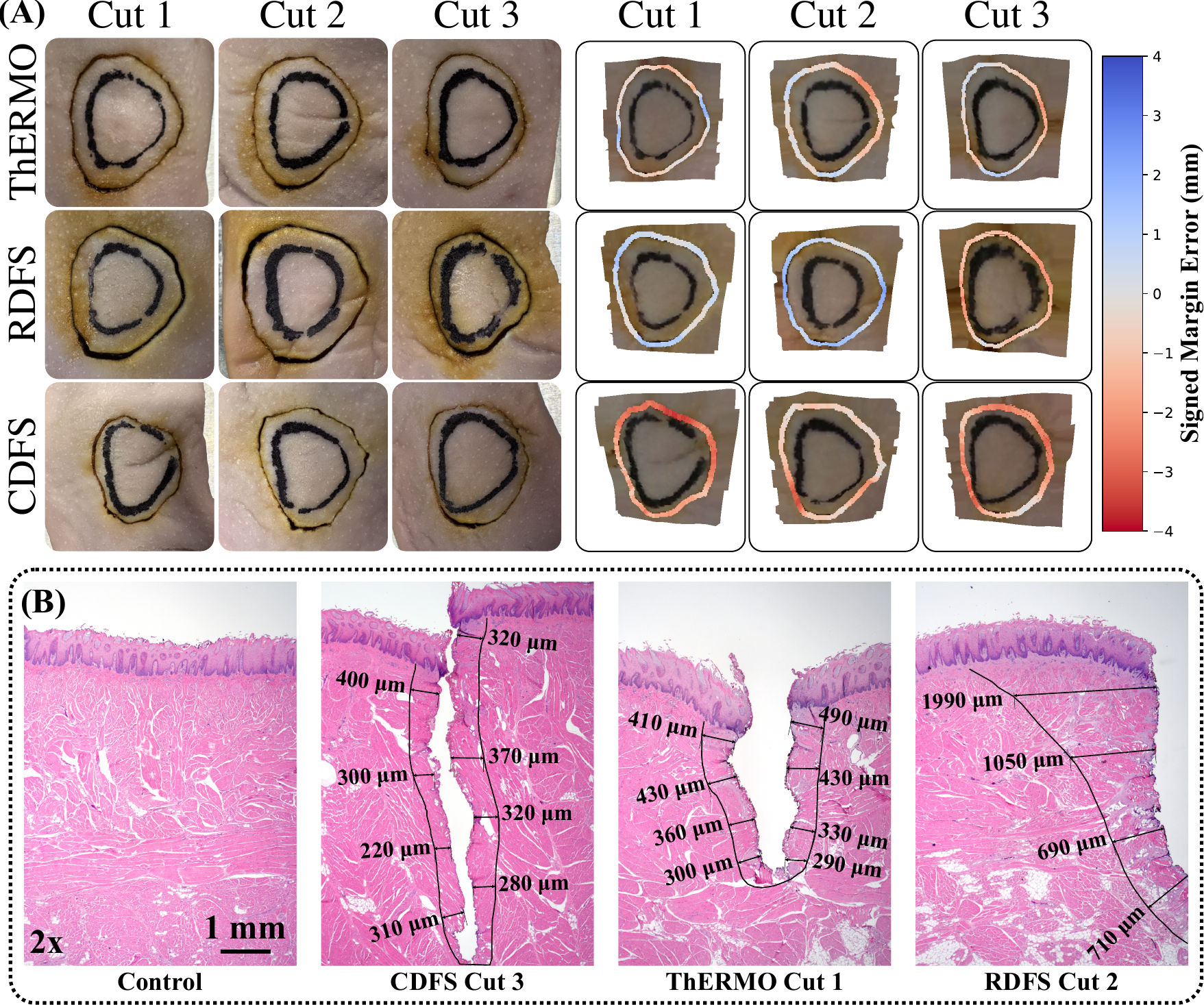}
    \caption{(A) A grid of photos shows each tongue sample immediately after cutting. Each cut is shown as a picture on the left and as a point-cloud scan on the right, overlayed with \acrfull{sme}. Each column represents one cutting method: \gls{rdfs} (\SI{2}{\milli\meter\per\second}), \acrfull{cdfs} (\SI{7}{\milli\meter\per\second}), and \gls{thermo}. The black path in the center of each tongue shows the virtual tumor boundary. The cut trajectory is \SI{5}{\milli\meter} from the center of the black edge. (B)  One representative \gls{he}-stained slide was selected for each method by the pathologist and is shown from least to most thermal injury. A black line denotes the extent of thermal injury, as measured by a pathologist. Eight measures are shown along this path. Due to the thermal injury width, only the left half of the cut is shown for \gls{rdfs}.}
    \label{fig:tongues}
\end{figure*}

\begin{table}
    \centering
    \small
    \caption{Summary of \acrfull{tie}, \acrfull{ame}, and \acrfull{sme} by cutting method, including estimated marginal means, pairwise contrasts of \gls{rdfs} (\SI{2}{\milli\meter\per\second}), \acrfull{cdfs} (\SI{7}{\milli\meter\per\second}), and \gls{thermo} from the \glspl{lmm}, and task completion time in seconds. }
    \label{tab:all_lmm}

    (a) Estimated Marginal Means \\[0.3em]
    \textit{Values are estimate $\pm$ SE. Best values are in bold.} \\[0.5em]
    \begin{tabular}{lccc}
        \toprule
        Method & AME (mm) & SME (mm) & TIE ($\mu$m) \\
        \midrule
        \gls{rdfs}    & $0.98 \pm 0.23$   & $\mathbf{0.08 \pm 0.45}$  & $515.43 \pm 64.34$ \\
        \gls{cdfs}    & $1.68 \pm 0.23$   & $-1.68 \pm 0.45$  & $293.68 \pm 58.62$ \\
        \gls{thermo} &   $\mathbf{0.69 \pm 0.23}$   & $-0.36 \pm 0.45$  & $\mathbf{246.57 \pm 70.68}$ \\
        \bottomrule
    \end{tabular}
    \vspace{1.0em}

    \setlength{\tabcolsep}{3.8pt}% tighten columns

    (b) Pairwise Contrasts from LMMs \\[0.3em]
    \textit{Values are estimate $\pm$\,SE.\;Significant contrasts ($\alpha=0.05$) are in bold.}\\[0.3em]
    \begin{tabular}{llcc}
      \toprule
      Metric & Contrast &
      Estimate $\pm$ SE &
      \(p\) \\
      \midrule
      \multirow{3}{*}{AME (mm)}
         & \gls{cdfs} – \gls{rdfs}                   & \(-0.71 \pm 0.32\)        & 0.071   \\
         & \textbf{\protect{\gls{cdfs} – \gls{thermo}}}  & \(\mathbf{-0.99 \pm 0.32}\) & \(\mathbf{0.022}\) \\
         & \gls{rdfs} – \gls{thermo}           & \(0.28 \pm 0.32\)         & 0.414   \\
      \midrule
      \multirow{3}{*}{SME (mm)}
         & \textbf{\protect{\gls{cdfs} – \gls{rdfs}}}          & \(\mathbf{1.76 \pm 0.64}\)  & \(\mathbf{0.033}\) \\
         & \gls{cdfs} – \gls{thermo}           & \(1.32 \pm 0.64\)          & 0.084   \\
         & \gls{rdfs} – \gls{thermo}           & \(0.44 \pm 0.64\)          & 0.519   \\
      \midrule
      \multirow{3}{*}{TIE ($\mu$m)}
         & \textbf{\protect{\gls{cdfs} – \gls{rdfs}}}          & \(\mathbf{221.75 \pm 47.04}\) & \(\mathbf{0.003}\) \\
         & \gls{cdfs} – \gls{thermo}           & \(-47.11 \pm 48.63\)       & 0.370   \\
         & \textbf{\protect{\gls{rdfs} – \gls{thermo}}}  & \(\mathbf{268.86 \pm 47.12}\) & \(\mathbf{0.001}\) \\
      \bottomrule
    \end{tabular}    
    \vspace{1.0em}

    (c) Task Completion Time per Trial \\ [0.3em]
    \begin{tabular}{lcccc}
        \toprule
        Method & Trial 1 & Trial 2 & Trial 3 & Mean \\
        \midrule
        \gls{rdfs} & \SI{53.68}{\second} & \SI{52.84}{\second} & \SI{50.44}{\second} & \SI{52.32}{\second} \\
        \gls{cdfs} & \SI{18.76}{\second} & \SI{19.28}{\second} & \SI{18.44}{\second} & \SI{18.83}{\second} \\
        \gls{thermo} & \SI{25.40}{\second} & \SI{20.12}{\second} & \SI{20.96}{\second} & \SI{22.16}{\second} \\
        \bottomrule
    \end{tabular}

\end{table}

A snapshot of the commanded speed, measured speed, measured deflection, and measured thermal width is shown in Fig.~\ref{fig:timeline}. Three timestamps are shown comparing an RGB image and a thermal image. 

\subsection*{Summary of Results}  

\gls{thermo} resulted in more accurate cuts relative to the intended surgical margin than \gls{cdfs}. Compared to \gls{rdfs}, \gls{thermo} also showed improved margin accuracy, though this difference did not reach statistical significance. 

For thermal injury, \gls{thermo} produced less injury to surrounding tissue than \gls{rdfs}, while the comparison with \gls{cdfs} did not show a statistically significant difference. In terms of procedure duration, \gls{rdfs} took longer to complete than both \gls{thermo} and \gls{cdfs}, which did not significantly differ from each other.

An expert reviewer rated the quality of resections from \gls{thermo} and \gls{rdfs} higher than those from \gls{cdfs}, which were downgraded due to poorer margin control. Overall, \gls{thermo} demonstrated advantages in margin accuracy and thermal safety compared to fixed-speed methods.

For a description of the main metrics and experimental conditions, please see Table~\ref{tab:cutting_methods_and_metrics}.

\subsection{Absolute Margin Error (AME)}

\begin{table}
    \centering
    \small
    \caption{Summary of cutting methods and outcome metrics used in the ex vivo tissue study.}
    \label{tab:cutting_methods_and_metrics}

    \textbf{(a) Cutting Methods} \\[0.3em]
    \begin{tabular}{lll}
        \toprule
        \textbf{Method} & \textbf{Speed} & \textbf{Description} \\
        \midrule
        \gls{rdfs} & \SI{2}{\milli\meter\per\second} & Speed used for \gls{astr}~\cite{ge_autonomous_2024}. \\
        \gls{cdfs} & \SI{7}{\milli\meter\per\second} & Expert surgeon's cutting speed~\cite{ge_enhancing_2024}. \\
        \gls{thermo} & Adaptive & Thermal-feedback optimized velocity. \\
        \bottomrule
    \end{tabular}

    \vspace{1.0em}

    \textbf{(b) Evaluation Metrics} \\[0.3em]
    \begin{tabular}{ll}
        \toprule
        \textbf{Metric} & \textbf{Description} \\
        \midrule
        \gls{sme} & Error between the measured and desired margin. \\
        \gls{ame} & Abs. error between the measured and desired margin. \\
        \gls{tie} & Avg. perpendicular extent of thermal injury. \\
        \bottomrule
    \end{tabular}
\end{table}

\begin{figure*}[tb!]
    \centering
        \includegraphics[width=\linewidth]{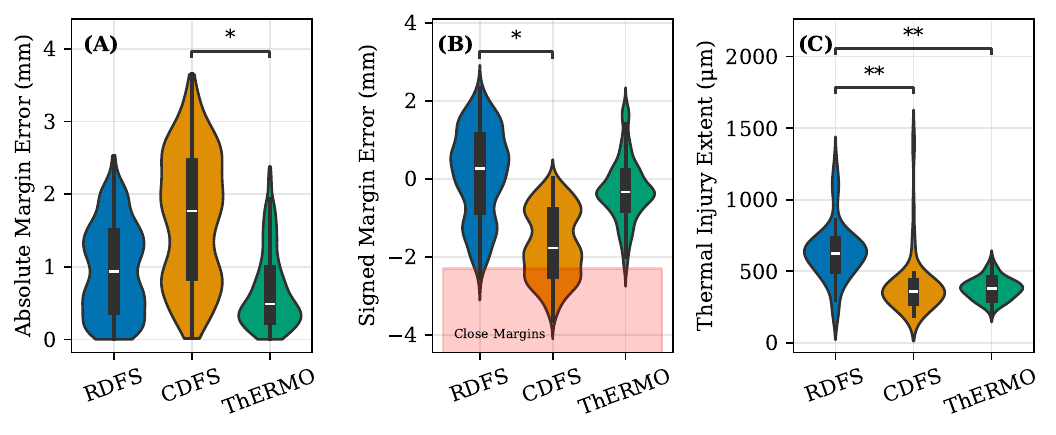}
        \caption{For all plots, ** refers to $p < 0.01$ and * refers to $p < 0.05$ for contrasts from the three \acrfullpl{lmm} ($\alpha=0.05$). (A)~\Acrfull{ame} is plotted in millimeters against cutting method ($n=3240$). \gls{thermo} vs. \gls{cdfs} showed a significant difference in \gls{ame} at $p<0.05$. (B)~\Acrfull{sme} is plotted in millimeters against cutting method ($n=3240$). Margins with \gls{sme} less than \SI{-2.2}{\milli\meter} (red region) are ``close margins"~\cite{li_margin_2019, zanoni_proposal_2017}. \gls{rdfs} vs. \gls{cdfs} were significantly different at $p<0.05$. (C)~\Acrfull{tie} is plotted in microns against cutting method ($n=216$). The difference between \gls{rdfs} vs \gls{thermo} and \gls{rdfs} vs \gls{cdfs} are significant at $p<0.01$.}
    \label{fig:box-plot}
\end{figure*}

\begin{figure}
    \centering
    \includegraphics[width=\linewidth]{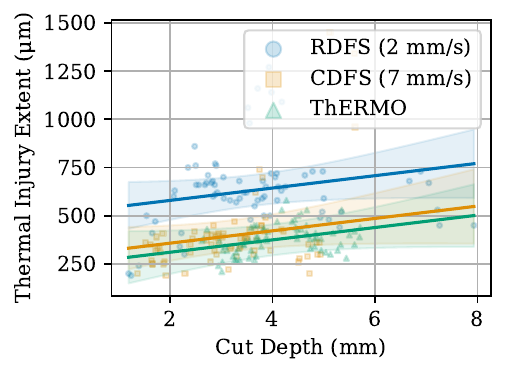}
    \caption{Cutting depth varied slightly during cutting. Statistical analysis revealed that depth was a significant predictor of \acrfull{tie} ($p = 0.034$). Cut depth impacted \gls{tie} at \SI{32.1}{\micro\metre\per\milli\metre}. The plotted lines show the \acrfull{lmm} fit with a 95\% confidence interval. \gls{thermo} produced the most accurate and most consistent cut depth (\SI{4.13}{\milli\meter}), followed by \gls{rdfs} (\SI{3.65}{\milli\meter}) and \gls{cdfs} (\SI{3.20}{\milli\meter})}
    \label{fig:depth_lmm}
\end{figure}

All nine tongue samples are shown in Fig.~\ref{fig:tongues}. A \gls{lmm} was fit to $n=3240$ margin accuracy measurements (360 samples per each of the 9 tongues) to estimate \gls{ame} from the ideal \SI{5}{\milli\meter} cutting path. Results are summarized in Tab.~\ref{tab:all_lmm}(a) and ~\ref{tab:all_lmm}(b) and distributions visualized in Fig.~\ref{fig:box-plot}A. The model revealed a significant effect of cutting method on \gls{ame} ($p < 0.05$). Estimated marginal means (±95\% CI) for \gls{ame} were \SI{1.68}{\milli\meter} [1.24, 2.13] for \gls{cdfs}, \SI{0.98}{\milli\meter} [0.53, 1.42] for \gls{rdfs}, and \SI{0.69}{\milli\meter} [0.25, 1.14] for \gls{thermo}. Pairwise comparisons of marginal means indicated that \gls{thermo} had significantly lower \gls{ame} than \gls{cdfs} ($p = 0.022$). The difference between \gls{rdfs} and \gls{cdfs} approached significance ($p = 0.071$). No significant difference was observed between \gls{rdfs} and \gls{thermo} ($p = 0.413$). 

\subsection{Signed Margin Error (SME)}
A second \gls{lmm} was fit to $n=3240$ margin accuracy measurements to estimate \gls{sme} from the ideal \SI{5}{\milli\meter} cutting path. Results are summarized in Tab.~\ref{tab:all_lmm}(a) and \ref{tab:all_lmm}(b) and visualized in Fig.~\ref{fig:box-plot}C. The model revealed a significant effect of cutting method on \gls{sme} ($p < 0.05$). Estimated marginal means (±95\% CI) for \gls{sme} were \SI{-1.68}{\milli\meter} [\SI{-2.57}{\milli\meter}, \SI{-0.80}{\milli\meter}] for \gls{cdfs}, \SI{0.08}{\milli\meter} [\SI{-1.07}{\milli\meter}, \SI{1.23}{\milli\meter}] for \gls{rdfs}, and \SI{-0.36}{\milli\meter} [\SI{-1.49}{\milli\meter}, \SI{0.77}{\milli\meter}] for \gls{thermo}. Pairwise comparisons of marginal means indicated that \gls{cdfs} produced significantly greater \gls{sme} than \gls{rdfs} ($p = 0.033$,) and have greater \gls{sme} compared to \gls{thermo}, though this difference was not statistically significant ($p = 0.084$). No significant difference was observed between \gls{rdfs} and \gls{thermo} ($p = 0.519$). 

Cutting depth varied modestly across cutting methods. All cuts were commanded to \SI{5}{\milli\meter} cut depth. \gls{thermo} produced the most accurate cut depth (\SI{4.13}{\milli\meter}), followed by \gls{rdfs} (\SI{3.65}{\milli\meter}) and \gls{cdfs} (\SI{3.20}{\milli\meter}). The impact of varying cutting depth is included in the statistical models for \gls{tie}.

\subsection{Thermal Injury Extent (TIE)}

Representative slides, selected by the expert pathologist, are shown in Fig.~\ref{fig:tongues}. A control sample is first depicted, then a black line denotes the thermal injury extent at 8 points along the cut profile, as measured by the pathologist. 

A \gls{lmm} fit to $n=216$ histological observations revealed significant differences among the three different cutting methods. A visualization of the \gls{tie} distributions is shown in Fig.~\ref{fig:box-plot}A. \gls{rdfs} had the greatest marginal mean \gls{tie} at \SI{515.43}{\micro\metre}, significantly exceeding both \gls{cdfs} (difference: \SI{221.75}{\micro\metre}, \(p=0.0033\))
and \gls{thermo} (difference: \SI{268.86}{\micro\metre}, \(p=0.0013\)). \gls{cdfs} (\SI{293.67}{\micro\metre}) did not differ significantly from \gls{thermo} (\SI{246.56}{\micro\metre}). Marginal means and contrasts are shown in Tab.~\ref{tab:all_lmm}(a) and \ref{tab:all_lmm}(b). Depth was a significant predictor of \gls{tie} (\(p=0.0342\)) with a predicted increase of \SI{32.1}{\micro\metre\per\milli\metre} of \gls{tie} per millimeter of depth. This trend is shown in Fig.~\ref{fig:depth_lmm}.

\subsection{Task Completion Time}

Task completion times are shown in Tab.~\ref{tab:all_lmm}c. \gls{cdfs} had the fastest completion time (\SI{18.83}{\second}). \gls{thermo} was 20.0\% slower than \gls{cdfs} at \SI{22.16}{\second}. Finally, \gls{rdfs} was slower than \gls{cdfs} by 178\% (\SI{52.32}{\second}) due to its low mean velocity. All methods were relatively consistent within the three trials and did not exhibit large variations in cutting time (\gls{rdfs}: 3.6\%, \gls{cdfs}: 2.4\%, \gls{thermo}: 14.5\%).

\subsection{Qualitative Clinical Assessment}

\gls{rdfs} received a score of 4/5, with the expert noting ``overall good speed, good accuracy in delineating circumferential margin''. \Gls{thermo} was also rated 4/5; the expert commented that it had ``good speed,'' but appeared ``more jagged on borders compared to method 1 [\gls{rdfs}],'' while offering the ``advantage of potentially less thermal damage''. \gls{cdfs} received a low score of 2/5. Although the expert mentioned its ``great speed'' and ``less thermal damage,'' they also noted ``much less accuracy in peripheral margin border compared to methods 1 [\gls{rdfs}] and 2 [\gls{thermo}]''.

% \end{document}

\section{Discussion}
\label{sec:discussion}
% \csname documentclass\endcsname[...]{subfiles}
% \graphicspath{{\subfix{../figures/}}}
% \begin{document}

This work demonstrates that incorporating closed-loop feedback, online estimation, and real-time optimization via \gls{thermo} yields substantial performance enhancements in autonomous ex vivo tumor resection. By reducing both \acrfull{tie} and \acrfull{ame} relative to fixed-velocity baselines, \gls{thermo} offers a more robust alternative to hand-tuned or predetermined velocity selection without the need for model training or tissue-specific calibration.

\subsection{Cut Accuracy}

Cut accuracy statistical analyses revealed that \gls{thermo} reduced \gls{ame} as compared to \gls{cdfs}. Elevated \gls{ame} is indicative of tissue snagging, a consequence of excessive cutting speed leading to reduced tissue denaturation. \gls{thermo}'s adaptive velocity profile was able to reduce snagging and improve \gls{ame} compared to both other methods, which is critical in tumor resection for preventing cancer recurrence while sparing healthy tissue. It also suggests that closed-loop feedback, regardless of the cutting speed, is necessary to ensure the most accurate cut possible. 

While \gls{thermo} reduced \gls{ame}, no statistical difference was found for \gls{sme}. \gls{rdfs} cuts trended further from the desired surgical margin, sacrificing extra healthy tissue. In contrast, \gls{thermo} and \gls{cdfs} both cut too close to the tumor, risking positive margins and recurrence. As shown in Fig.~\ref{fig:box-plot}B, \gls{cdfs} has the largest proportion of ``close margins", highlighting a significant risk of worse post-operative outcomes~\cite{li_margin_2019, zanoni_proposal_2017}.

Variation in cut depth compared to the commanded depth of \SI{5}{\milli\meter} was observed across all methods, potentially indicating some inaccuracy in executing the cut plan. Increased cutting depth may cause excess thermal damage, while decreased cutting depth may lead to insufficient cutting. Variation towards shallower cutting may be due to deflection in the tool and tissue when cutting. Deeper cutting depths may be due to thermal conduction from the bottom of the needle, desiccating tissue below the tool as well. Inaccuracy in general may be due to error accumulation in the robot's kinematic chain, or simply tissue deformation that occurred when preparing the tissue slides for microscopy. Variations in cut depth also change the thermal energy deposited into the tissue, which may impact open-loop cutting methods that assume a specific constant thermal flux. Cutting depth may also impact \gls{sme} and \gls{ame}, though complete depth measurements were not available with the accuracy measures. The potential impact on accuracy and the correlation between cutting depth and \gls{tie} underscore the need for closed-loop cut depth monitoring in future studies.

\subsection{Thermal Injury}

\gls{thermo} achieved significantly lower thermal injury than \gls{rdfs} and performed comparably to \gls{cdfs}. These findings suggest that \gls{thermo}'s velocity adaptation modulates thermal energy deposition, maintaining cut accuracy while minimizing thermal injury. The thermal injury seen through microscopy also verifies that measured temperatures from the T3 thermal camera indicate real structural injury in the tissue; specifically, the observed connective tissue bluing indicates collagen denaturation.

Although \gls{thermo} had the highest average cut depth, it still maintained lower \gls{tie} than \gls{rdfs}, highlighting the benefit of adaptive feedback in adjusting for variations in cutting depth or tissue properties. Notably, \gls{thermo} also reduced peak \gls{tie} (Fig.~\ref{fig:box-plot}C) compared to the other methods, suggesting that though \gls{cdfs} generates less thermal energy into the tissue, feedback-based velocity augmentation is more effective at adapting to changing cut characteristics and thus maintaining consistent results. The blinded pathologist review provides evidence linking measured thermal width to cellular thermal injury.   

\subsection{Task Completion Time}

In the operating room, task completion time is critical to minimizing costs, reducing patient anesthesia time, and preventing complications~\cite{jackson_does_2011}. Due to its increased speed, \gls{thermo} performed similarly to \gls{cdfs} in cutting time, substantially faster than \gls{rdfs}. \gls{thermo}'s ability to achieve similar cut accuracy performance to \gls{rdfs} without sacrificing operating time is critical to optimal operating room performance.

\subsection{Expert Analysis}

Expert analysis showed that \gls{thermo} and \gls{rdfs} were both effective and provided good cut quality, while \gls{cdfs} did not perform well. \gls{rdfs} was noted to have the best margins of the three methods, but with comparatively greater thermal injury. This feedback may indicate that clinicians value margin accuracy more highly than thermal injury, which may be used to inform future cost function weights. The expert noted that \gls{cdfs} was inadequate due to its insufficient tumor margin and jagged cutting edges, caused by tissue snagging. However, the expert did note that cutting speed for \gls{cdfs} seemed most appropriate, perhaps since it aligns best with the surgeon's measured cutting speed. The increased speed did result in lowered visible thermal injury, but this benefit was not enough to increase the overall score. 

While \gls{thermo} produced comparable expert scores to \gls{rdfs}, the noted jaggedness may be due to high-velocity trajectories along regions of high curvature, causing tool orientation discontinuities. Future work should evaluate kinematic smoothing strategies or a shorter electrosurgical tool to reduce orientation change artifact, especially since jagged cuts may reduce surgeon confidence in system reliability and increase velocity command error.

\subsection{Limitations}

This study was performed on ex vivo porcine tissue, which lacks the perfusion, temperature, and mechanical properties of living human tissue. As a result, both thermal injury and margin deformation may not exactly translate to human studies. Further in vivo studies are needed to optimize clinical translation. Additionally, while \gls{thermo} reduced thermal injury, the observed extent may be lower than would be expected in real clinical scenarios, as the tissue used here was cut at room temperature, not typical human body temperatures, and other indicators of damage, like inflammation or fibrosis that may have occurred in living tissue, were absent. Increasing the ambient temperature of the tissue will reduce the margin for thermal damage, likely requiring increased cut velocities, but these effects may be negligible due to the exponential nature of temperature increase. Blood perfusion may have complex effects, however, as it both removes heat from the direct vicinity and spreads the heat to a larger area.
 
The relationship between local cutting velocity and thermal injury was not explicitly studied but warrants future analysis across a broader range of velocities. Finally, while \gls{he} staining provided thermal injury assessment under an expert's review, enzyme-based methods such as \gls{nbt} or \gls{ldh} may better highlight thermal damage, though they may perform best on freshly resected in vivo tissue~\cite{gibson_improving_2017, ewertowska_histological_2024}. Alternatively, at high temperatures (\SIrange{69}{75}{\celsius}), collagen loses its birefringence and can therefore indicate thermal injury by use of polarizing filters on \gls{he}-stained slides~\cite{pearce_irreversible_2018}.

Tissue with higher curvature will require more dexterous robotic manipulation to appropriately reach around the tumor margin. This added dexterity may challenge the fixing of the thermal camera to the end effector, but efforts towards thermal camera miniaturization are promising. Furthermore, only one tissue variety is tested here. Different organs have different tissue properties and should be independently verified to further show generalization. 

% \end{document}

\section{Conclusions and Future Works}
% \csname documentclass\endcsname[...]{subfiles}\graphicspath{{\subfix{../figures/}}}
% \begin{document}

\Gls{thermo} significantly reduced \gls{tie} and \gls{ame} in ex vivo tumor resection relative to fixed-speed baselines, demonstrating that thermal-feedback and optimization-based control may improve clinical outcomes in surgery without requiring tuning or data-driven training. \gls{thermo} can simultaneously improve both results without the tradeoffs that fixed velocity methods demonstrate, supporting the potential for clinical translation.

While this study focused on comparisons to other autonomous methods, a comparison to human-operated robotic surgery could further contextualize \gls{thermo}'s performance. Prior works have shown that robotic cutting causes worse thermal damage than humans, but with \gls{thermo}, this may no longer be true~\cite{ewertowska_histological_2024}. A study with more fixed velocities could refine the relationship between velocity and thermal injury. 

Unlike surgeons, \gls{thermo} does not ever raise the tool or move backwards. Adding these behaviors to the controller may further improve performance. This will require a re-framing of the optimization problem to include multiple actions. Optimization and parameter estimation can be improved by utilizing more accurate dynamics models. To do so would require re-framing the problem using model-predictive control frameworks, wherein a cost function is minimized over a short time horizon using a predictive model.

\gls{thermo}'s utility with other control variables such as depth of cut or electrosurgical power may also be evaluated. Though only currently applied to autonomy, novice surgeons may benefit from \gls{thermo} as a training tool and surgical safety monitor.

Finally, improving thermal camera miniaturizing and integrating 3D point clouds fusion are key technological steps towards clinical translation. Ultra-low pixel pitch technology~\cite{lutz_towards_2018} or fiber optics must be implemented to improve existing resolution limits. As thermal sensing technology advances, thermal feedback will become increasingly practical for real-time control in surgery.

% \end{document}

% \section*{Acknowledgements}

\bibliographystyle{IEEEtran}
\bibliography{references.bib}

\vfill\null

\end{document}